\documentclass[sigconf]{acmart}

\def\BibTeX{{\rm B\kern-.05em{\sc i\kern-.025em b}\kern-.08emT\kern-.1667em\lower.7ex\hbox{E}\kern-.125emX}}
    
\copyrightyear{2019} 
\acmYear{2019} 
\setcopyright{acmcopyright}
\acmConference[SIGIR '19]{Proceedings of the 42nd International ACM SIGIR Conference on Research and Development in Information Retrieval}{July 21--25, 2019}{Paris, France}
\acmBooktitle{Proceedings of the 42nd International ACM SIGIR Conference on Research and Development in Information Retrieval (SIGIR '19), July 21--25, 2019, Paris, France}
\acmPrice{15.00}
\acmDOI{10.1145/3331184.3331320}
\acmISBN{978-1-4503-6172-9/19/07}

\begin{document}

%
\title{Learning Unsupervised Semantic Document Representation for Fine-grained Aspect-based Sentiment Analysis}

%

\author{Hao-Ming Fu}
\affiliation{\institution{National Taiwan University}}
\email{r06922092@ntu.edu.tw}

\author{Pu-Jen Cheng}
\affiliation{\institution{National Taiwan University}}
\email{pjcheng@csie.ntu.edu.tw}

%
\renewcommand{\shortauthors}{}

%
\begin{abstract}
Document representation is the core of many NLP tasks on machine understanding. A general representation learned in an unsupervised manner reserves generality and can be used for various applications. In practice, sentiment analysis (SA) has been a challenging task that is regarded to be deeply semantic-related and is often used to assess general representations. Existing methods on unsupervised document representation learning can be separated into two families: sequential ones, which explicitly take the ordering of words into consideration, and non-sequential ones, which do not explicitly do so. However, both of them suffer from their own weaknesses. In this paper, we propose a model that overcomes difficulties encountered by both families of methods. Experiments show that our model outperforms state-of-the-art methods on popular SA datasets and a fine-grained aspect-based SA by a large margin.
\end{abstract}

%
%


%

%

%
\maketitle

\section{Introduction}
An informative document representation is the key to many NLP applications such as document retrieval, ranking, classification and summarization. Learning without supervision reserves generality of learned representation and takes advantage of large corpus with no labels.

There are two families on learning document representation without supervision: Sequential and non-sequential models. The former takes ordering of words into consideration when processing a document, often with sequential architectures such as RNN. The effectiveness of these models drops significantly when the text being processed gets much longer than a sentence. Consequently, simpler models from non-sequential family often outperforms sequential ones on the task. However, semantic meaning is intuitively lost when ordering of words is discarded.

For instance, consider these two reviews on beer: `` I love the smell of it, but the taste is terrible.'' and ``This one tastes perfect, but not its smell.'' Obviously, for models discarding the order of words, recognizing which aspect each sentimental word ``love'', ``terrible'', ``perfect'', ``not'' refers to is not possible.

The overall sentiment of the reviews cannot be well captured either without aspect separation. That is because an overall sentiment can be viewed as a combination of individual aspects weighted by their importance. The best a non-sequential model can do with a mixture of sentimental words without knowing importance of each of them is a rough average. 

In this paper, we propose a model that overcomes difficulties encountered by both sequential and non-sequential models. Our model is tested on widely used IMDB \cite{IMDB} sentiment analysis dataset and the challenging aspect-based Beeradvocate \cite{Beeradvocate} dataset. Our results significantly outperform state-of-the-art methods on both datasets.

\section{Related works}
Non-sequential methods range widely from early Bag-of-Word model and topic models including LDA to more complex models such as Denoising Autoencoders \cite{DEA}, Paragraph Vectors\cite{doc2vec} and doc2vecC\cite{doc2vecC}. Sequential methods emerge quickly in recent years thanks to the development of neural networks. Models for text sequence representation include Skip-thoughts \cite{skip-thought}, a sentence level extension from word level skip-gram model, and many other CNN or RNN based methods.
    
    Modeling a document as a group of sentences is not a new idea, but an effective design to learn without supervision under this framework is yet to be done.	The closest work to our model should be doc2vecC and Skip-thoughts Vectors. Our model is similar to doc2vecC in the way that our model represents a document by averaging embedding of sentences in it, while doc2vecC averages embedding of words in the document. Besides, both doc2vecC and our model explicitly use mean of embedding during training to assure a meaningful aggregation of embedding vectors. Our model is similar to Skip-thought Vectors in the way that both models try to capture relations between adjacent sentences. Skip-thought Vectors chooses a generic prediction model, while our model projects sentences into a shared hidden space and learn meaningful features by managing relations of sentences in the space.

\section{The proposed model}
Given a document \(D\) composed of \(n\) sentences \([s_0, s_1, \ldots, s_n]\) in order, our goal is to obtain a vector representation \(v_D\) for the document. Note that \([\ldots]\) stands for an ordered list in the rest of this paper.

\subsection{Overview}
    Figure 1 is an overview of our model. The purpose of the model is to obtain a vector representation for document \(D\) in an unsupervised manner. We update variables in the model by training it to predict a target sentence among some candidate sentences given its context sentences. The context sentences are defined by \(k\) sentences on each side of the target sentence \(s_t\). Namely, \(S_{cntx}=[s_{t-k},\ldots, s_{t-1}, s_{t+1},\ldots, s_{t+k}]\).
    
    Besides the target sentence, \(r\) negative samples are coupled with each target sentence \(s_t\). The model will calculate a probability distribution over these \(r+1\) candidate sentences to make prediction. We refer to the list of candidate sentences as \(S_{cdd}=[s_t,s_{neg_1},\ldots,s_{neg_r}]\). The model will output \(r+1\) scalars, corresponding to each sentence in \(S_{cdd}\). These scalars are referred to as logits of the sentences. A higher logit indicates a higher probability is distributed to the sentence by the model. Logit of the target sentence \(s_t\) is denoted as \(l_t\) and logits of negative samples \(s_{neg_1},\ldots,s_{neg_r}\)are denoted as \(l_{neg_1},\ldots,l_{neg_r}\).
    
     According to Chen et al.\cite{neg-samp}, with those logits given, optimizing the following loss function will approximate optimizing the probability distribution over all possible sentences in the world:
     
     \begin{equation}
loss=-log(\sigma(l_t))+\sum_{i=0}^r log(\sigma(l_{neg_i}))\end{equation}
    
Applying negative sampling, a softmax function is not literally operated while a distribution over infinite number of all possible sentences in the world is optimized. After the model is trained this way, it can be used to calculate a vector representation for a document.

    \begin{figure}[h]
  \centering
  \includegraphics[width=\linewidth]{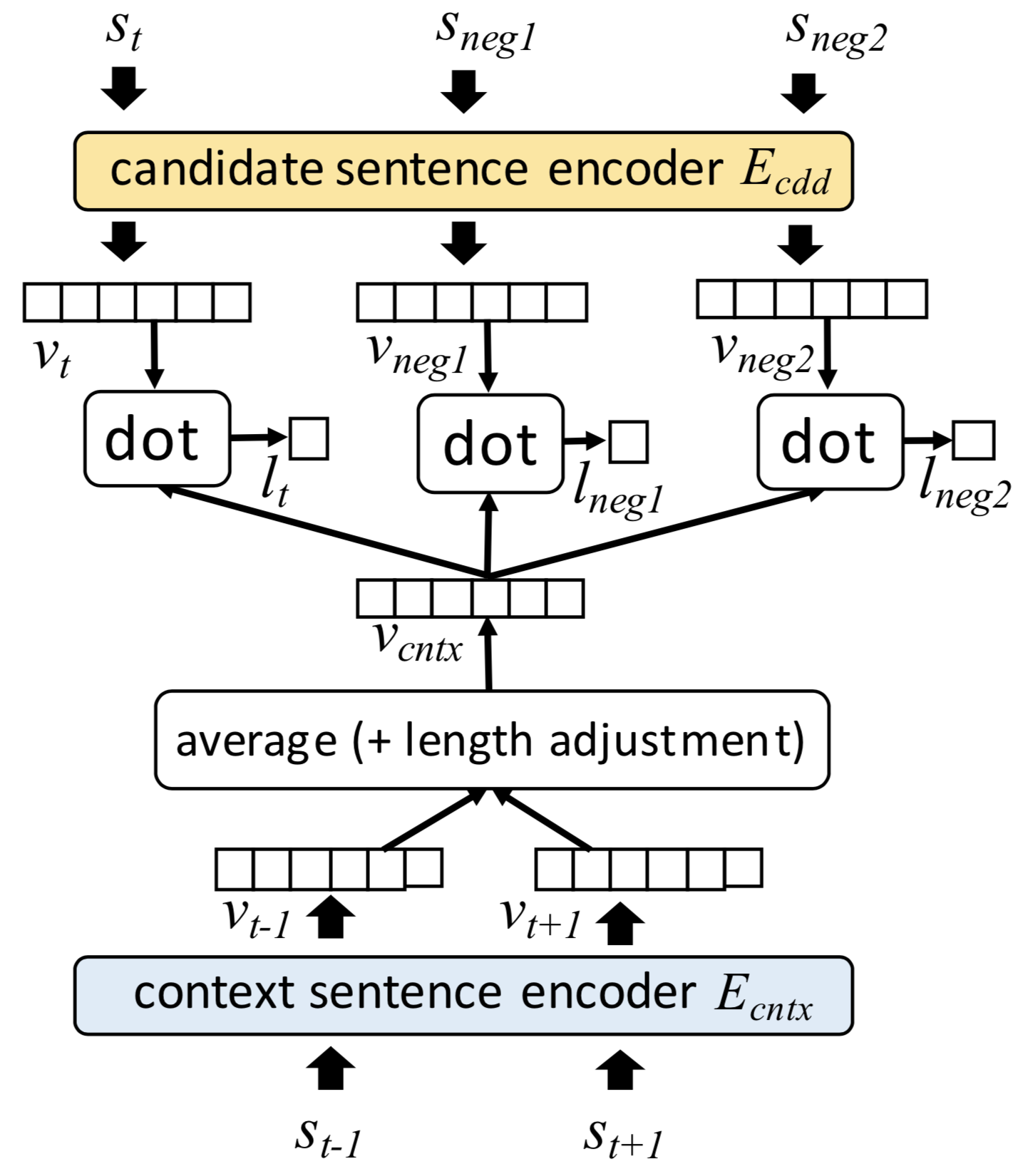}
  \caption{Overview of our model. In the figure, number of context sentences on each side is 1 and number of negative samples \(r\) is 2. Context sentences \(s_{t-1}, s_{t+1}\)are fed to the model from the bottom. The target sentence \(s_t\) and negative samples \(s_{neg_1}, s_{neg_2}\)are fed from the top. Logit of the target sentence \(l_t\)and negative samples \(l_{neg_1}, l_{neg_2}\)are obtained in the middle. These will be used to calculate the loss.}
  \Description{.}
\end{figure}

\subsection{Architecture}

\subsubsection{model}
As illustrated in Figure 1, we use sentence encoders to encode a sentence into a fixed-length sentence vector. Two sentence encoders are used in the model, the context encoder \(E_{cntx}\) and the candidate encoder \(E_{cdd}\). Sentences in \(S_{cntx}\)are encoded into sentence vectors \(V_{cntx} = [v_{t-k}, \ldots, v_{t-1}, v_{t+1}, \ldots, v_{t+k}]\) by \(E_{cntx}\). Those in \(S_{cdd}\) are encoded into a target sentence vector \(v_t\) and negative samples vectors \(V_{neg} = [v_{neg_1}, \ldots, v_{neg_r}]\) by \(E_{cdd}\). To merge information captured by each sentence vector in \(V_{cntx}\) into a single context vector, vectors in \(V_{cntx}\) are element-wise averaged. The obtained context vector is called \(v_{cntx}\).

\(v_{cntx}\) will go through a process called length adjustment except when calculating \(L_{cntx}\) in Section \(3.3.1\). Length adjustment process will normalize \(v_{cntx}\) and lengthen it to the average length of sentence vectors which are used to obtain \(v_{cntx}\) itself. The process is as follow:
\begin{equation}
adjusted \quad v_{cntx} = \frac{v_{cntx}} {length(v_{cntx})} \times \frac{\sum_{v_i \in V_{cntx}}length(v_i)}{size(V_{cntx})}
\end{equation}
where \(length(x)\) denotes \(l2\) norm of \(x\) and \(size(y)\) denotes number of elements in \(y\). This process solves the length vanishing problem of element-wise averaging many vectors.

Now, we have a single vector \(v_{cntx}\)containing unified information from context sentences. If the sentence vector of a candidate is similar to \(v_{cntx}\), it is probability the sentence to be predicted. Similarity is evaluated with inner product. So, \(v_{cntx}\)will dot with the target sentence vector \(v_{t}\) and negative sentence vectors in \(V_{neg}\) to obtain a logit for each of them. Logit of the target sentence is called \(l_{t} = dot(v_{cntx}, v_{t})\) and logits of negative samples are called \(l_{neg_1}, \ldots, l_{neg_r}\), where \(l_{neg_i} = dot(v_{cntx}, v_{neg_i})\).

With these logits, the loss can be calculated with Equation (1).

\begin{table}
  \caption{Structure of sentence encoders. For consistency with Figure 1, first layer is placed at the bottom and the last layer at the top.}
  \label{tab:freq}
  \begin{tabular}{ccl}
    \toprule
    Layer type & parameters\\
    \midrule
    \it{Output Layer} & \it{a fixed-length sentence vector.}\\
    Dropout & dropout rate 0.5\\
    Fully connected & 100 nodes\\
    Fully connected & 1024 nodes with ReLU\\
    Global average pooling &\\
    Max pooling & size 2 with stride 2\\
    Convolutional & 256 filters with size 2\\
    Convolutional & 256 filters with size 2\\
    Max pooling & size 2 with stride 2\\
    Convolutional & 256 filters with size 2\\
    Convolutional & 128 filters with size 2\\
    Word embedding table & embedding dimension 100\\
    \it{Input Layer} & \it{a sentence.}\\
    \bottomrule
\end{tabular}
\end{table}

\subsubsection{Sentence encoders}
\(E_{cntx}\) and \(E_{cdd}\) have the same structure, as elaborated in Table 1. Nevertheless, they do not share variables except the word embedding table. This allows a sentence to be represented differently when playing different roles. We choose convolutional networks for sentence encoders for its simplicity and efficiency of training. Note that a global average pooling layer is placed on top of convolutional layers to form a fix-length vector for sentences of variable length. 

\subsection{Training}

	During training, a list of sentences \(S_D = [s_0, s_1, \ldots, s_n]\) from a single document \(D\)  is fed to the model as a single training sample. The total loss to be minimized, \(L_{total}\), is the weighted sum of two terms: the context loss \(L_{cntx}\) and the document loss \(L_{doc}\). The model is then trained end to end by minimizing \(L_{total}\).

\subsubsection{Context loss}

	For each sentence in \(S_D\), \(k\) sentences before and \(k\) sentences after the target sentence are given in \(S_{cntx}\) as context sentences. Besides this, randomly selected negative samples \(s_{neg_1},\ldots,s_{neg_r}\) are selected from sentences in other documents in the dataset. Length adjustment process is not applied when calculating context loss. Target sentence logit \(l_t\) and negative sentences logits \(l_{neg_1}, l_{neg_2}, \ldots, l_{neg_r}\) are obtained and used to calculate \(L_{cntx_t}\)with Equation (1). The context loss \(L_{cntx}\) is defined by averaging losses from each sentence in \(S_D\) except the first \(k\) and the last \(k\) sentences for incomplete context sentences.
\begin{equation}
L_{cntx} = \frac{\sum_{i=k+1}^{n-k}L_{cntx_i}}{n-2k}
\end{equation}
	where \(L_{cntx_i}\) is the context loss of a single target sentence.

\subsubsection{Document loss}

	For document loss, there are only two differences from context loss: 1) length adjustment process is applied on \(v_{cntx}\). 2) all the sentences in \(S_D\), including the target sentence \(s_t\) itself, are regarded as context sentences for each target sentence. Consequently, each sentence in \(S_D\) can be used as target sentence.
The document loss \(L_{doc}\) is defined by averaging losses from all the sentences in the document:
\begin{equation}
L_{doc} = \frac{\sum_{i=1}^{n}L_{doc_i}}{n}
\end{equation}
\subsubsection{Total loss}
	The total loss is the weighted sum of context loss and document loss. A hyper-parameter \(\alpha\) is used to assign weights. Total loss \(L_{total}\) is obtained by:
\begin{equation}
L_{total} = \alpha \times L_{cntx} + (1-\alpha) \times L_{doc}
\end{equation}
\(L_{total}\) is then minimized to update model variables. In particular, \(L_{cntx}\) and \(L_{doc}\)are responsible for capturing local and global relations among sentences respectively. \(L_{doc}\) also guarantees an effective aggregation for sentence vectors.

\subsection{Inference of document representation}
For a document \(D\), its representation is the length adjusted average of sentence vectors from all sentences in it. No extra training is needed for new documents seen for the first time. Notice that it is exactly the context vector \(v_{cntx}\)used for calculating \(L_{doc}\). It is explicitly used during model training on purpose. This leads the model to learn sentence vectors that can be effectively aggregated by average. Also, the aggregated representation is guaranteed to be informative since it is also learned during training.

\section{Experiments}

We first test our model on the widely used IMDB review dataset \cite{IMDB} for SA. To go further, we test our model on the Beeradvocate beer review dataset \cite{Beeradvocate} for aspect-based SA. This dataset challenges document representations with much more fine-grained SA.

\subsection{Sentiment analysis}

\subsubsection{Dataset}

We use IMDB review dataset in this sentiment analysis experiment. The dataset consists of 100k movie reviews. 25k of the data are labeled for training and another 25k are labeled for testing. The rest 50k reviews are unlabeled. Both training and testing data are balanced, containing equivalent number of reviews labeled as semantically positive and negative.

\subsubsection{Experiment design}

We follow the design of Chen\cite{doc2vecC} to assess our model under two settings: use all available data for representation learning or exclude testing data. Both of them make sense since representation learning is totally unsupervised. After model training, a linear SVM classifier is used to classify learned document representation under supervision. The performance of the classifier, evaluated by accuracy, indicates the quality of learned representation.

We compare our model with intuitive baseline methods including Bag-of-Words, Word2Vec+AVG and Word2Vec+IDF, word-embedding based method like SIF \cite{SIF}, sequential models including RNN language model, Skip-thought Vectors \cite{skip-thought} and WME \cite{WME}, and non-sequential models including Denoising Autoencoder \cite{DEA}, Paragraph Vectors \cite{doc2vec} and Doc2vecC \cite{doc2vecC}. Representative models from both sequential and non-sequential families along with some intuitive baselines are compared with.

We use a shared word embedding table of 100 dimensions and train it from scratch. Dimensions of learned document representation are set as 100, which can be inferred from the outputs of sentence encoders. Dropout rate is 0.5 and \(\alpha\) is tuned to be 0.7.

\begin{table}
\caption{Sentiment analysis results on IMDB dataset in accuracy (\%). Extra adv. column marks extra advantages out of experiment settings. D for representation dimension greater than 100, E for external data other than IMDB dataset used, S for supervision by label during training. Methods in the sequential family are marked with (Seq.). Results sources: \cite{WME} for WME, \cite{SIF} for SIF and \cite{doc2vecC} for others.}
  \begin{tabular}{cccc}
    \toprule
    Methods&Extra adv.&Acc.(\%) &Acc.(w/o test,\%) \\
    \midrule
    Skip-thought Vectors (Seq.)&  D, E & -& 82.58\\
    SIF with GloVe& E & - &  85.00 \\
    RNN-LM (Seq.)&S& 86.41 & 86.41\\
    Word2Vec + AVG  &E& 87.89& 87.31\\
    Bag-of-Words & D & 87.47 & 87.41\\
    Denoising Autoencoders& -  & 88.42 & 87.46\\
    Paragraph Vectors & - & 89.19& 87.90\\
    Word2Vec + IDF  &E& 88.72& 88.08\\
    Doc2VecC & - & 89.52& 88.30\\
    WME (Seq.)& E & - & 88.50 \\
    \bottomrule
    Our model (Seq.) & - & \textbf{92.78} &\textbf{90.83}\\
\end{tabular}
\end{table}

\subsubsection{Results and discussion}

The results are shown in Table 2. Our model considerably outperforms state-of-the-art models. As we discussed, sequential models suffer from long text and non-sequential models lose semantic information for discarding ordering of words. Our model, on the other hand, successfully overcomes the difficulties encountered by both families of methods. Our model considers ordering of words within each single sentence, which is considered the fundamental unit of a concept. At the same time, instead of processing long text at once, pieces of concepts extracted from sentences are effectively aggregated to form a meaningful representation of documents.

\subsection{Aspect-based sentiment analysis}

Aspect-based sentiment analysis is a more challenging task for document representation. Besides capturing an overall image of a document, detailed information mentioned in only part of the document has to be recognized and well preserved. We test the ability of our model to learn a single representation that includes information from all different aspects. We compare our model with doc2vecC on this task, since it is the strongest competitor in the sentiment analysis experiment without any extra advantage.

\subsubsection{Dataset and Experiment design}

We choose the Beeradvocate beer review dataset for aspect-based SA task. It consists of over 1.5 million beer reviews; each has four aspect-based scores and one overall score. All the scores are in the range of 0 to 5 and given by the reviewers. The four aspects are appearance, aroma, palate and taste. For a fair comparison with the SA experiment, we only use the first 500k reviews of the dataset.

To follow the settings of the SA experiment, we reassign labels to each aspect to simplify it to a binary classification task. A review is labeled as positive/negative on a certain aspect if its score on the aspect is not lower/higher than 4.5/3.5. For each aspect, we construct two pools of positive and negative reviews respectively. We randomly select 50k samples from each pool. The selected data are split in half for training and testing. Now we have 50k balanced data for training and 50k data for testing for each aspect.

In this experiment, all available data (500k data used in the experiment) are used for representation learning. For each aspect, a linear SVM classifier will be trained. We use the same parameters as on IMDB review dataset.

\subsubsection{Results and discussion}
\begin{table}
  \caption{Results of aspect-based sentiment analysis on Beeradvocate dataset. Reported numbers are accuracy (\%).}
  \label{tab:freq}
  \begin{tabular}{cccccc}
    \toprule
Model & Appearance&Aroma&Palate&Taste & Overall\\
\midrule
doc2vecC & 80.826&82.810&82.500&86.154 & 82.366\\
\bottomrule
Our model & \textbf{85.070}&\textbf{86.695}&\textbf{86.795}&\textbf{91.020 }&\textbf{ 87.280}\\
\end{tabular}
\end{table}

Results of the experiment are shown in Table 3. Our model far outperforms doc2vecC on every aspect-based classification tasks including overall. The results indicate that information of all aspects is better captured and stored in a single vector learned by our model. It also illustrates the generality of our model to perform well on different aspects and tasks with different difficulties.

We notice that even though doc2vecC does not explicitly consider ordering of words, it still achieves an acceptable accuracy on aspect-based classification. This may be caused by the fact that many words used in the reviews are aspect-related on its own. For instance, ``delicious'' is a strongly taste-related word that is useful for aspect-based sentiment analysis even without knowing its context.

Surprisingly, we find in experiments that performance of our model is hardly sensitive to any of the hyper-parameters except \(\alpha\). We tuned \(\alpha\) in the range between 0 and 1 and picked 0.7. We find the value generalizable to different tasks and datasets. As for other hyper-parameters, we find the model insensitive to them in a wide range. That is why we use exactly the same parameters on both IMDB and Beeradvocate datasets. This observation indicates the effectiveness as well as robustness of our model design.

\section{Conclusions}

Experimental results show that our model outperforms state-of-the-art unsupervised document representation learning methods by a large margin on both classic SA task and its aspect-based variance.

We attribute this improvement to the design of our model that enables it to reserve ordering of words and aggregate sentence vectors effectively at the same time. Splitting long text into sentences avoids the curse of length for sequential models. Aggregation with average is made effective by explicitly using the obtained representation during training.

%
\bibliographystyle{ACM-Reference-Format}
\bibliography{sample-base}

\end{document}